\documentclass[3p,authoryear]{elsarticle}
\usepackage{natbib}
\usepackage{fullpage}
\usepackage[english]{babel}
\usepackage[utf8]{inputenc}

\usepackage{amsfonts}
\usepackage{amsmath}
\usepackage{mathtools}
\usepackage{graphicx}
\usepackage{subcaption}
\usepackage{hyperref}
\usepackage{multirow,booktabs}
\usepackage{float,subcaption,overpic}
\usepackage[colorinlistoftodos]{todonotes}
\usepackage{rotating}
\usepackage{algorithm,algpseudocode,caption}
\usepackage{wrapfig}
\usepackage{hyperref}

\usepackage{etoolbox}
\usepackage{tikz}

\newcommand{\W}{{\bf W}}
\newcommand{\C}{{\bf C}}

\title{\textbf{Bracketing brackets with bras and kets}\vspace{-.05in}}

\author[uwphysics]{Emily Clark\corref{cor}} \ead{eclark7@uw.edu}
  \cortext[cor]{Corresponding author. \emph{E-mail:} eclark7@uw.edu} 
    \author[boeing]{Angelie Vincent}
  \author[uwamath]{J. Nathan Kutz}  
  \author[uwme]{and Steven L. Brunton}
    \address[uwphysics]{Department of
    Physics, University of Washington, Seattle, WA 98195,
    United States}
        \address[boeing]{The Boeing Company, Seattle, WA 98008}
  \address[uwamath]{Department of
    Applied Mathematics, University of Washington, Seattle, WA 98195,
    United States}
    \address[uwme]{Department of
    Mechanical Engineering, University of Washington, Seattle, WA 98195,
    United States\vspace{-.35in}}
      
\date{}

\begin{document}

\begin{abstract}
Brackets are an essential component in aircraft manufacture and design, joining parts together, supporting weight, holding wires, and strengthening joints. 
Hundreds or thousands of unique brackets are used in every aircraft, but manufacturing a large number of distinct brackets is inefficient and expensive. 
Fortunately, many so-called ``different" brackets are in fact very similar or even identical to each other. 
In this manuscript, we present a data-driven framework for constructing a comparatively small group of representative brackets from a large catalog of current brackets, based on hierarchical clustering of bracket data. 
We find that for a modern commercial aircraft, the full set of brackets can be reduced by 30\% while still describing half of the test set sufficiently accurately. 
This approach is based on designing an inner product that quantifies a multi-objective similarity between two brackets, which are the ``bra" and the ``ket" of the inner product.  
Although we demonstrate this algorithm to reduce the number of brackets in aerospace manufacturing, it may be generally applied to any large-scale component standardization effort.  
\end{abstract}


\maketitle

\vspace{-.2in}
\section{Introduction}
\label{sec:introduction}

Component part standardization has been shown to be highly beneficial in manufacturing processes, leading to higher efficiency and reduced costs \citep{collier1981measurement, collier1982aggregate, brownell1990budgetary, perera1999component, tsubone1994component, vakharia1996operating}. In the aircraft manufacturing industry, Boeing already employs a platform approach for the overall design of many aircraft, in which new models may be rapidly designed to fit specific marketplace niches by rescaling existing parts or groups of parts \citep{meyera2001perspective}. 
However, some component parts in the manufacturing process have yet to be standardized, including brackets, which join parts, strengthen joints, and hold wires. Currently, a new bracket is designed for nearly every new purpose. Reducing the number of unique brackets would result in improved efficiency. However, standardizing the set of brackets is a complex problem and must account for the dimensions of the joint or intersection, the type of interface, the frequency of use, and cost of materials and manufacturing, where these variables can be more or less important on a case by case basis. In this work, we present an optimization strategy for reducing a large set of brackets down to a comparatively small set of standardized brackets, based on hierarchical clustering of the bracket data.  We find that by using our machine learning strategy, the number of unique brackets can be reduced by approximately 30\% with little reduction in performance. 

Industry standardization is often achieved by building a family of standardized component parts, where the parts can be combined to design new products. It is highly inefficient to custom design each component part for every new product, so having a predetermined set of possible parts to draw from can lead to significant savings in both time and money, as well as guaranteeing that components will be compatible with one another. A prominent example of product family design is in the automotive industry \citep{muffatto1999introducing}, where many companies utilize platforms, a standardized group of parts such as the floor, suspension, engine, and fuel tank, which serves as the base for multiple products. Unique features such as the car body and steering can be added to the platform, yielding an entirely new product far more efficiently in terms of engineering and manufacturing than if it were custom designed.

The parts that make up the platform and the unique parts that are added to it are often specifically engineered for that purpose.  
However, what about the case where there is a large library of parts from past products that have similar functionalities? It may be beneficial to standardize the library into a reduced set. In fact, \cite{perera1999component} performed an analysis of the savings accrued through component part standardization and found that costs are reduced throughout the entire life cycle of a product. These savings occur during product development, where engineering efforts are reduced; manufacturing, as materials can be purchased more cheaply in bulk, and machine setup and labor costs are reduced; distribution, since taking inventory is simpler, and facilities are more robust to damaged components; usage, where the customer has better access to spare parts and maintenance support; and disposal, with recycling being easier with fewer unique parts. The one major disadvantage sited is that if parts are too standardized, they may not be able to meet customers' satisfaction completely.


Despite the clear benefits of component part standardization, the literature on how to standardize a large library of past parts is surprisingly sparse, with most sources on standardization focusing on its benefits, as in \cite{perera1999component, vakharia1996operating} or quantification of standardization, e.g. \cite{collier1981measurement, wacker1986component}. \cite{junmin2007similarity} describe a standardization method based on calculating the topological similarity of parts' form features (geometrical shapes that make up the part), but this method is very general, as opposed to the bracket standardization problem, where we can leverage the fact that every bracket of the same type has the same underlying structure.

However, in recent years big data and machine learning techniques have proven highly effective for predictive manufacturing, which improves efficiency, quality, and cost savings; see, e.g.,~\cite{harding2006data, wang2007applying, lee2013recent, lechevalier2014towards, esmaeilian2016evolution}. Moreover, dimensionality reduction methods have been applied to manufacturing in many contexts:~\cite{guo2017identification} use topological data analysis (TDA) for prediction and fault detection and~\cite{guo2018sparse} extend TDA through sparse sampling. Principal component analysis (PCA) has been widely used to predict assembly variation, for example,~\cite{manohar2018predicting, camelio2002compliant, carnelio2006identification, lindau2013statistical}. And~\cite{zhang2006comparative} adopt PCA for surface characterization and outlier rejection. These methods prove that machine learning and dimensionality reduction techniques can improve prediction, control, and decision making for manufacturing, and now we employ a common unsupervised learning algorithm for the reduction of a large set of bracket data.


Our approach to standardization is to consider past bracket data and perform clustering in parameter space to identify a relatively small number of brackets to approximately represent the entire set (a similar framework to that of coresets \citep{badoiu2002approximate}). The part standardization problem can be viewed in several possible ways, but we identify two main questions:
\begin{enumerate}
\item Given a set number of standardized brackets to be designed from the training set, how accurately are the brackets in the test set represented?
\label{question1}

\item How many standardized brackets are required for a given accuracy in representing the test set?
\label{question2}
\end{enumerate}

There are also multiple ways to quantify accuracy. Accuracy could be taken as the mean error of the test set, where the error is the distance between a test bracket and the nearest representative bracket, for some distance metric. Or one could set a distance threshold and count the number of test set brackets that fall below it, i.e. are said to be correctly described by a standardized bracket. Thus, there are two mathematically rigorous ways to formulate questions 1 and 2, as described in Section \ref{subsec:problem}.

The goal of bracket standardization is to identify a standardized bracket that most closely matches the specification of a gap or joint along with the required tolerances. While the number of standardized brackets is predetermined, since accuracy is of primary importance in aircraft manufacture, if the closest standardized bracket is not within tolerance, the system will be allowed to return a new bracket that fits the gap or joint perfectly.

To this end, we use hierarchical clustering \citep{ward1963hierarchical}, an unsupervised machine learning algorithm offering flexibility in both within- and between-cluster distance metrics, to group the brackets in the training set by similarity, and choose one bracket from each cluster as a representative. We leverage the manufacturing process to achieve higher accuracy: to manufacture a bracket, it is extruded or cut and bent into shape, and then the hole pattern is stamped into it. Therefore, we cluster on geometrical and hole pattern variables separately. We then apply the set of standardized brackets to the test set and calculate the accuracy both in terms of error and number correctly categorized. By varying the number of clusters, we balance accuracy and a reduction in the number of unique brackets. We leave room for user-specified tolerances when calculating the number of brackets correctly categorized. 

\begin{figure}
\centering
\includegraphics[width=\textwidth]{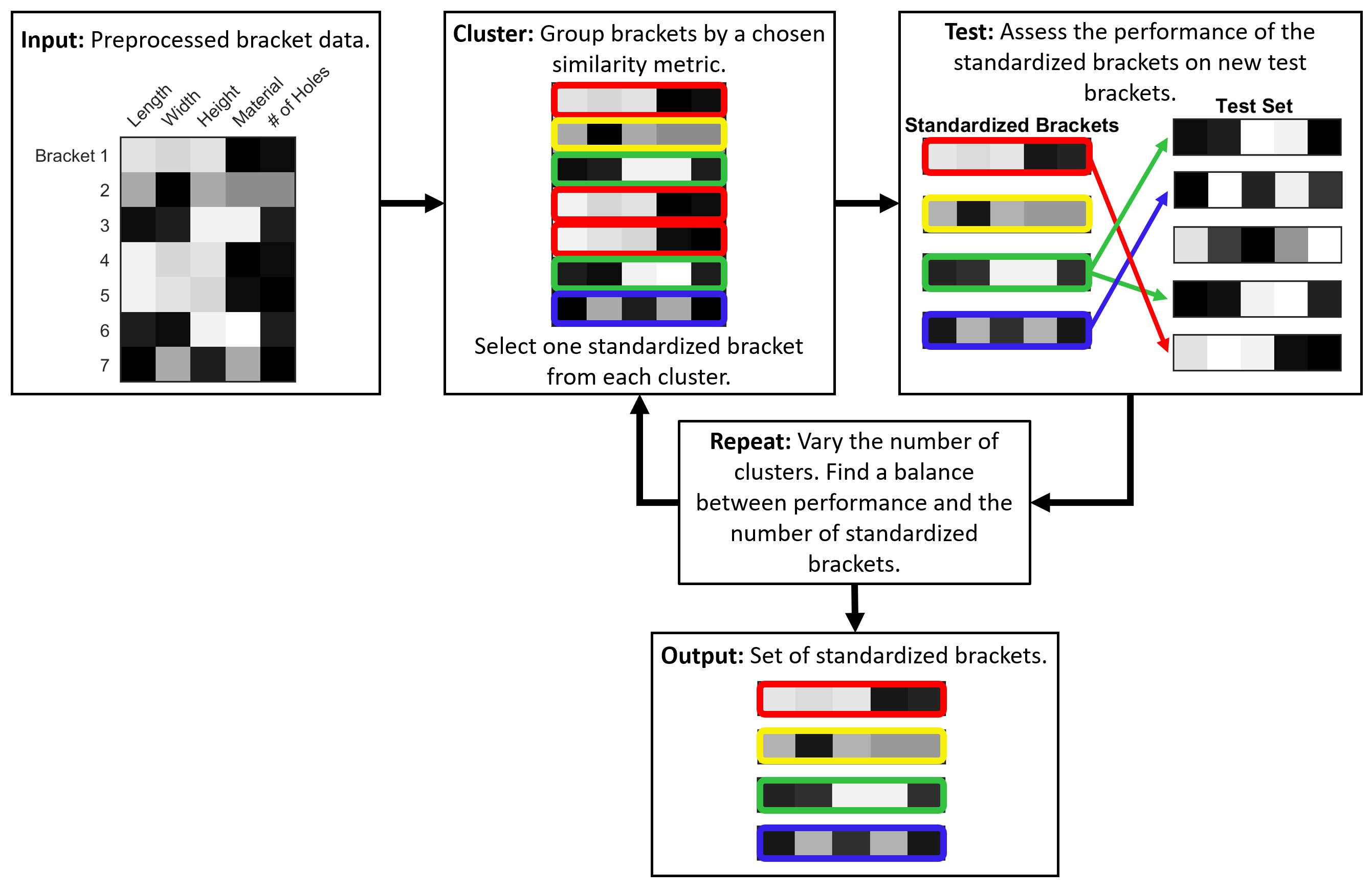}
\caption{The basic bracket standardization procedure.}
\label{fig:flowchart}
\end{figure}

An outline of our procedure is given in Figure \ref{fig:flowchart}. 
In Section \ref{sec:methods}, we state the optimization problems, present the bracket data set, and provide some background on hierarchical clustering. In Section \ref{sec:results}, we show the results of our bracket clustering algorithm, and we present conclusions in Section \ref{sec:conclusions}.

\section{Methods} \label{sec:methods}

\subsection{Problem statement} \label{subsec:problem}
Consider a set $\mathcal{S}_N$ of standardized brackets ${\bf S}_j$, $j = 1,2,\dots, N$ and a set of test brackets $\boldsymbol{\sigma}_i$, $i = 1,2,\dots,m$. Define a distance metric $d_{ij}$ to calculate errors between a test bracket $\boldsymbol{\sigma}_i$ and standardized bracket ${\bf S}_j$. We choose the $L_2$ norm, $d_{ij}=\sqrt{\left\langle {\bf S}_j - \boldsymbol{\sigma}_i | {\bf S}_j - \boldsymbol{\sigma}_i \right\rangle}$; in general, we will choose the standardized bracket ${\bf S}_j$ that is closest to test bracket $\boldsymbol{\sigma}_i$, resulting in the smallest distance $d_{ij}$. Let $\C \in\mathbb{R}^m$ be a vector whose $i^{\text{th}}$ entry is 1 if the error of test bracket $\boldsymbol{\sigma}_i$ is less than some chosen tolerance $\tau$, and zero if the error is greater than $\tau$. Finally, select a total error tolerance $\mathcal{E}$ and a total number of test brackets $T$ that must be categorized correctly. 
Then, for a set number $N$ of standardized brackets, question 1 from the introduction can be written as follows:

\begin{subequations}
\begin{align}
\min_{\mathcal{S}_N} \sum_{i=1}^m \sqrt{\left\langle {\bf S}_j - \boldsymbol{\sigma}_i | {\bf S}_j - \boldsymbol{\sigma}_i \right\rangle},
\label{equation1a}
\end{align}
\begin{align}
\max_{\mathcal{S}_N} ||\C||_0, \hspace{6pt} \text{where } 
\left\{\begin{array}{c c}
C_i = 1, & \sqrt{\left\langle {\bf S}_j - \boldsymbol{\sigma}_i | {\bf S}_j - \boldsymbol{\sigma}_i \right\rangle} \le \tau,\\
\vspace{-6pt}\\
C_i = 0, & \sqrt{\left\langle {\bf S}_j - \boldsymbol{\sigma}_i | {\bf S}_j - \boldsymbol{\sigma}_i \right\rangle} > \tau.
\end{array}\right.
\label{equation1b}
\end{align}
\end{subequations}
Equation~\ref{equation1a} seeks to minimize the $L_2$ error at a given number of standardized brackets, and Equation~\ref{equation1b} seeks to maximize the number of test set brackets that fall below the error threshold.

The second question from the introduction can be written as the following optimization problems:
\begin{subequations}
\begin{align}
\min N \hspace{6pt} \text{s.t. } \min_{\mathcal{S}_N} \sum_{i=1}^m \sqrt{\left\langle {\bf S}_j - \boldsymbol{\sigma}_i | {\bf S}_j - \boldsymbol{\sigma}_i \right\rangle} < \mathcal{E},
\label{equation2a}
\end{align}
\begin{align}
\min N \hspace{6pt} \text{s.t. } ||\C||_0 \ge T, \hspace{6pt} \text{where } 
\left\{\begin{array}{c c}
C_i = 1, & \sqrt{\left\langle {\bf S}_j - \boldsymbol{\sigma}_i | {\bf S}_j - \boldsymbol{\sigma}_i \right\rangle} \le \tau,\\
\vspace{-6pt}\\
C_i = 0, & \sqrt{\left\langle {\bf S}_j - \boldsymbol{\sigma}_i | {\bf S}_j - \boldsymbol{\sigma}_i \right\rangle} > \tau.
\end{array}\right.
\label{equation2b}
\end{align}
\end{subequations}
In other words, find the smallest number of standardized brackets such that either the $L_2$ error is below a given value (Equation~\ref{equation2a}), or the desired number of test set brackets have errors below a certain threshold (Equation~\ref{equation2b}). Note that while $\tau$ is written as a threshold on the overall error, one can instead select a different tolerance for each variable in parameter space, and require that $\C_i = 1$ only if every variable of bracket $\boldsymbol{\sigma}_i$ is within its tolerance.

Our approach to addressing the above optimization problems will be to perform unsupervised clustering and vary the number of clusters $N$ (standardized brackets). We then test the performance with differing numbers of standardized brackets, and look for the value of $N$ that is simultaneously optimal in minimizing the number of standardized brackets and maximizing the performance, also see Figure~\ref{fig:flowchart}.

There are several other possible error metrics, including the $L_1$ norm, the $L_\infty$ norm, and a weighted norm $\sqrt{\left\langle {\bf S}_j - \boldsymbol{\sigma}_i | {\bf S}_j - \boldsymbol{\sigma}_i \right\rangle}_\W$, where $\W$ is a diagonal matrix of weights. The latter error metric is potentially useful to the bracket standardization problem, as it would allow us to emphasize the variables that are more important or difficult to get right during clustering or selection. However, we find that it does not improve clustering for our purposes when compared to the $L_2$ norm, and we choose to weight the variables during selection by setting individual variable thresholds, since this is ultimately more interpretable and practical for aircraft manufacturing.

\subsection{Data and preprocessing} \label{subsec:data}

\begin{figure}
\centering
\includegraphics[width=\textwidth]{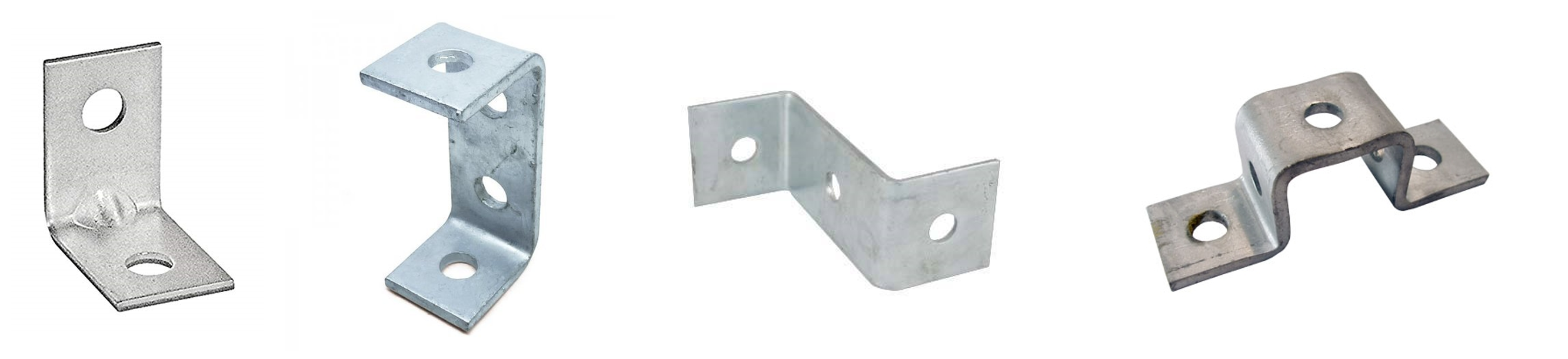}
\caption{A basic example of each category of bracket. From left to right, these are: angle~\cite{angleamazon}, C or U~\cite{cindiamart}, Z~\cite{zindiamart}, and hat brackets~\cite{hatmeteor}.}
\label{fig:bracket_examples}
\end{figure}

This dataset will consider a set of 5207 brackets from a modern commercial aircraft. There are four primary types of brackets in the data set, with a general example of each pictured in Figure \ref{fig:bracket_examples}. Angle brackets are the most numerous, with 1963 samples; there are also 774 Z brackets, 413 C or U brackets, and 172 hat brackets. The remaining brackets are either of unknown type or are flat.

There are a total of 44 parameters describing each bracket, comprising bracket identification number, type, total length, width, and depth, minimum and maximum thickness, number of fastener groups, number of segments, total number of holes, and maximum hole diameter. For each segment there are also length, thickness, angle with respect to horizontal, number of fasteners, maximum fastener diameter, minimum fastener separation in the extrusion direction, and minimum fastener separation in the segment length direction.

The data set requires some preprocessing. The first step is to separate the brackets by type: in this work, we will only consider angle brackets, as they are the most numerous type. We then discard the bracket identification number, as it does not factor into clustering, and set all blank entries to zero. We perform one-hot encoding on the integer variables such as the number of fastener groups, which extends the total number of variables to 74. Finally, we rescale the columns of the resulting data matrix to have unit variance, to ensure that all parameters are treated equally in clustering.

\subsection{Hierarchical clustering} \label{subsec:clustering}

\begin{figure}
\centering
\includegraphics[width = \textwidth]{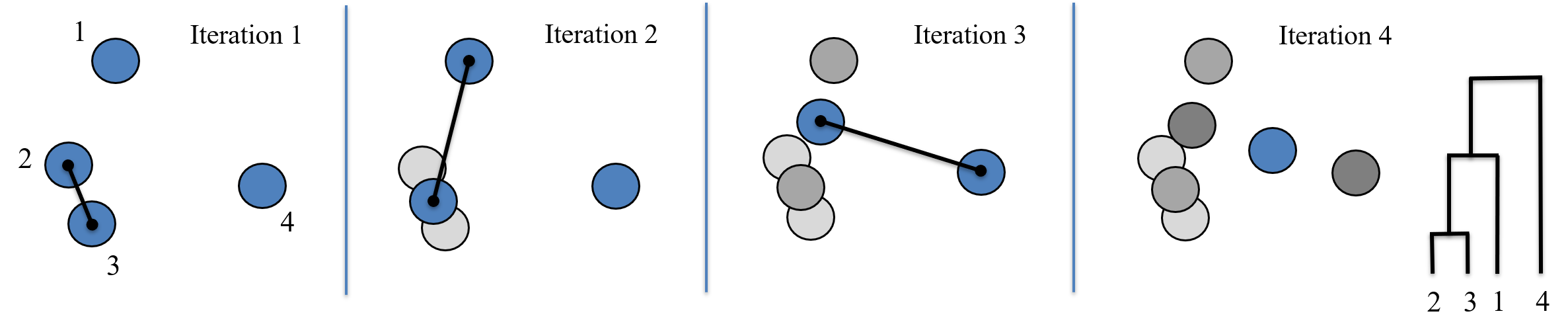}
\vspace{-12pt}
\caption{An illustration of hierarchical clustering on four data points, adapted from \cite{Brunton2019book}. Here, the distance metric is the euclidean distance, and at every iteration the next cluster's location is the average location of the two points or clusters it contains. Note that the length of the branches on the dendrogram is proportional to the distance between the points.}
\label{fig:clustering}
\end{figure}

Hierarchical clustering, introduced in \cite{ward1963hierarchical}, is a method of iteratively grouping data points by similarity. It is one of the canonical unsupervised machine learning techniques, extracting groupings from unlabelled data while remaining agnostic to the final number of clusters, allowing for user input but providing metrics to aid in choosing the number of clusters if desired. The algorithm is straightforward. At the first iteration, the pairwise distances between every data point are calculated using some chosen distance metric, and the closest two data points are merged into a single cluster. In subsequent iterations, the procedure is repeated, where the distance between clusters, or linkage, is calculated with a predetermined function of the data points within the clusters. The results of the clustering are usually presented in a dendrogram. See Figure \ref{fig:clustering} for a simple illustration of the clustering procedure.

For a distance metric, we choose the Euclidean or $L_2$ norm,
\begin{align}
d_{ij} = \sqrt{\left\langle \boldsymbol{\sigma}_i - \boldsymbol{\sigma}_j | \boldsymbol{\sigma}_i - \boldsymbol{\sigma}_j \right\rangle},
\end{align}
for two brackets, $\boldsymbol{\sigma}_i$ and $\boldsymbol{\sigma}_j$. We employ Ward's linkage, or the incremental sum of squares, to calculate the within-cluster distance:
\begin{align}
D_{pq} = \sqrt{\frac{2n_pn_q}{n_p+n_q}} \left\langle {\bf c}_p - {\bf c}_q | {\bf c}_p - {\bf c}_q \right\rangle,
\end{align}
for clusters $p$ and $q$ containing $n_p$ and $n_q$ elements, with centroids ${\bf c}_p$ and ${\bf c}_q$.

The Euclidean norm provides an unbiased distance metric in the high-dimensional parameter space, while Ward's linkage is a measure of the increase in the within-cluster sum of squares produced by joining two clusters. It reduces to the Euclidean norm in the case of two singleton clusters, and we find that it produces the best performance results for this data set when compared to other possible linkages.

\section{Results}
\label{sec:results}

We perform clustering on the 1963 angle brackets, breaking them up into a training set of 1563 randomly-chosen brackets and reserving the remaining 400 for the test set. We average over 50 cross validations.

For an error measure, we take the unweighted $L_2$ norm,
\begin{equation}E_i = \sqrt{\left\langle {\bf S}_j - \boldsymbol{\sigma}_i | {\bf S}_j - \boldsymbol{\sigma_i} \right\rangle},
\label{eq:error}
\end{equation}
where $E_i$ is the error for test bracket $\boldsymbol{\sigma}_i$, and ${\bf S}_j$ is the standardized bracket that is closest to $\boldsymbol{\sigma}_i$. We choose ${\bf S}_j$, the training set bracket that represents cluster $j$, by finding the member of the cluster with the lowest error between it and the other brackets in cluster $j$.

To calculate the error in the $k^{\text{th}}$ variable alone, calculate
\begin{equation}
E_{ik} = \sqrt{\left\langle S_{jk} - \sigma_{ik} | S_{jk} - \sigma_{ik} \right\rangle}.
\end{equation}

\begin{figure}
\centering
\includegraphics[width=\textwidth]{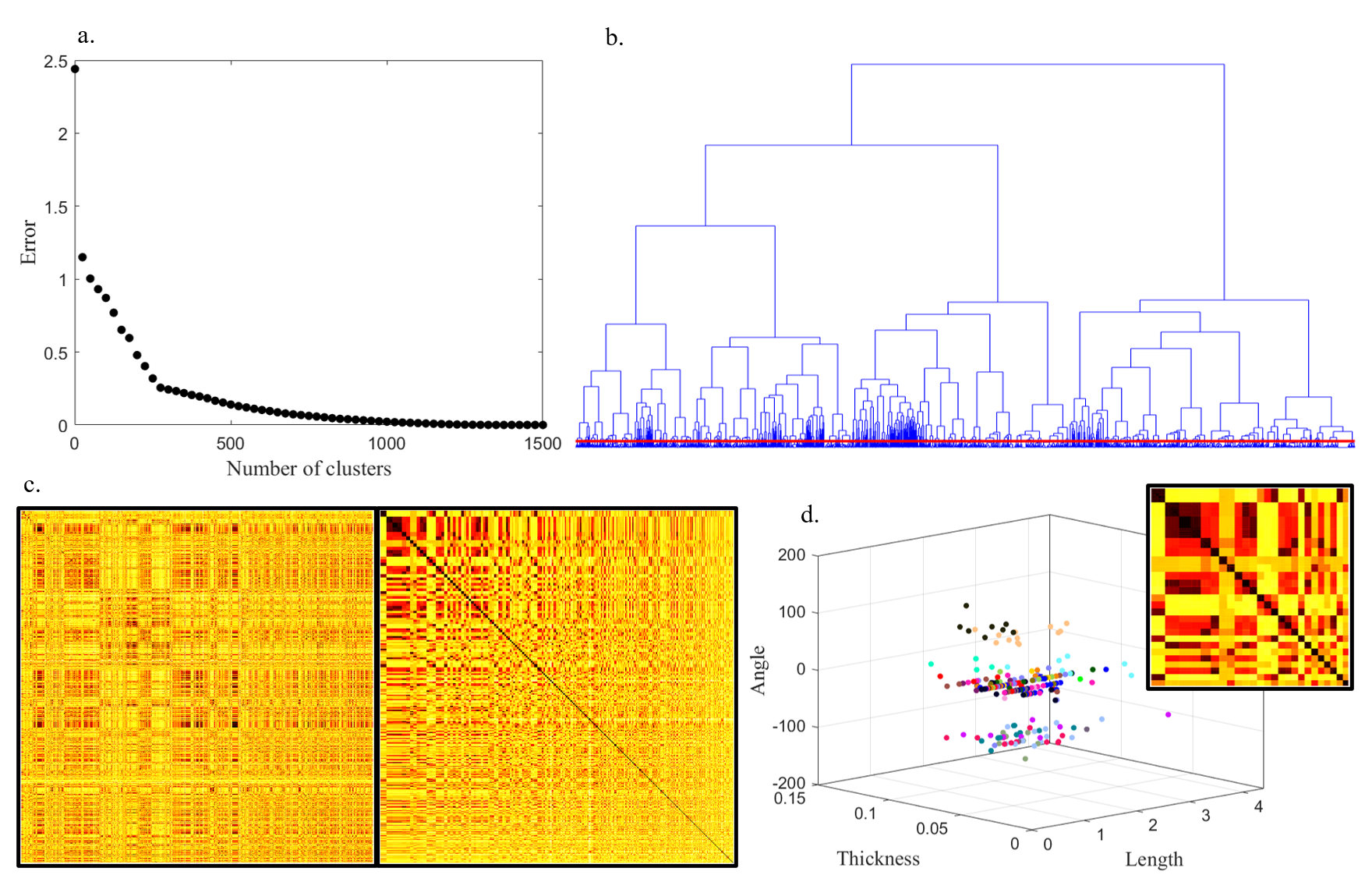}
\caption{Some initial clustering results. Subplot a. gives the average within-cluster error versus the number of clusters. Subplot b. gives the clustering dendrogram, with the red horizontal line representing the threshold above which there are 400 clusters. Subplot c. shows two distance matrices, where the left plot was generated with the unsorted data set, and on the right, the brackets have been sorted into 400 clusters, organized by cluster size. Finally, subplot d. plots Segment 1 length, thickness, and angle for the first 25 out of 400 clusters, as data points colored by cluster. The inset is a close-up of the sorted distance matrix, showing the 477 brackets that are included in those first 25 clusters.}
\label{fig:ClusterResults}
\end{figure}

We begin with some example results of the hierarchical clustering on the entire data set of 1963 brackets in Figure~\ref{fig:ClusterResults}. The upper left plot shows the average within-cluster error, as the number of clusters is increased from 1 to 1500. This average error is of order $10^{-3}$ by 1200 clusters and continues to decrease as the number of clusters is increased. However, the errors are order 1 at a small number of clusters, indicating that while there is some structure in the data and the number of unique brackets needed can be reduced, drastic reductions may not be possible. Interestingly, there is an elbow in the plot at 275 clusters, beyond which the slope of the error line decreases sharply, perhaps indicating an optimal number of standardized brackets. Unfortunately, this elbow disappears when clustering on a smaller training set, as shown in later figures.

The plot on the upper right of Figure~\ref{fig:ClusterResults} shows the dendrogram from clustering on the full data set, with a line above which there are 400 clusters. The $y$-axis shows the error between clusters. Subplot c. shows two distance matrices, where the first is for the unsorted data set, and the second shows the data sorted into 400 clusters, in order of cluster size. It is evident that the majority of the clusters contain only one or two brackets, though the largest contains 34; the low within-cluster error is apparent for all groups. The plot on the bottom right shows the length, thickness, and angle of segment 1 for the brackets in the first 25 out of 400 clusters (comprising 477 brackets, about $1/4$ of the data set), colored by cluster. The groupings are apparent in these three variables, though not perfect. For example, most gold data points have angles around $90^\circ$, but a few have angles near zero. Finally, the inset shows the distance matrix for the 477 brackets in the first 25 clusters. Though the brackets in the first cluster have very low errors with each other and high errors with brackets outside the cluster, the discrimination between the next few clusters is less clear, suggesting that it should be possible to combine them all into one large cluster with little loss of performance. This is supported by subplot a., which shows relatively little decrease in error from 275 to 400 clusters, though it should be noted that the errors are still fairly large in this region, as demonstrated by subplot d.

\begin{figure}
\centering
\includegraphics[width = 0.5\textwidth]{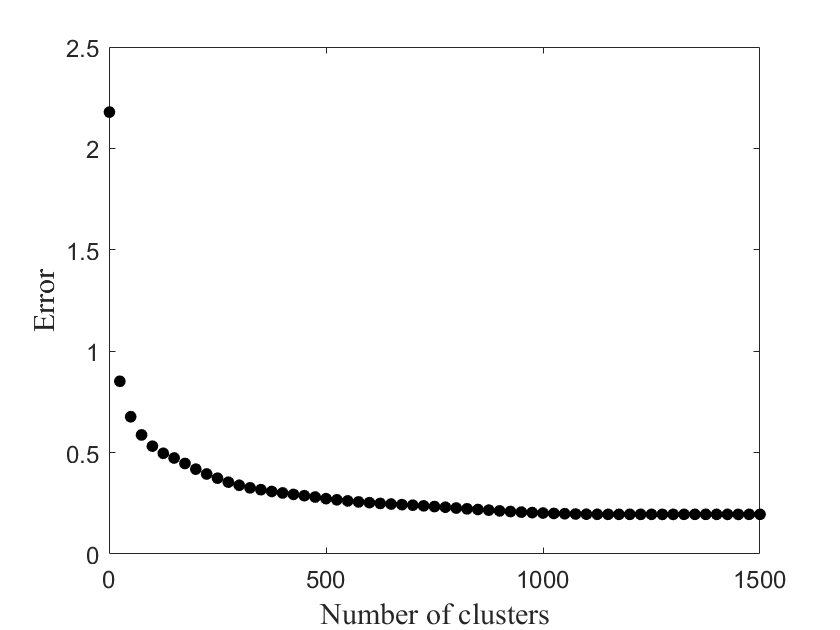}
\caption{The average test set error as a function of the number of clusters.}
\label{fig:TestError_vs_Num}
\end{figure}

We move on to testing how well the standardized brackets can describe a test set. Figure \ref{fig:TestError_vs_Num} gives the average test set error, given by Equation~\ref{eq:error}, as the number of clusters is increased from 1 to 1500. The results are encouraging, with a Pareto point at about 200 standardized brackets, just 13\% of the test set, with an error of approximately 0.4. This suggests that there is little performance improvement from adding more standardized brackets beyond the Pareto point, but since the minimum error reached is 0.19 at 1500 clusters, it is likely that many test set brackets are not able to be represented well enough by the nearest standardized bracket.

\begin{figure}
\centering
\includegraphics[width = 0.9\textwidth]{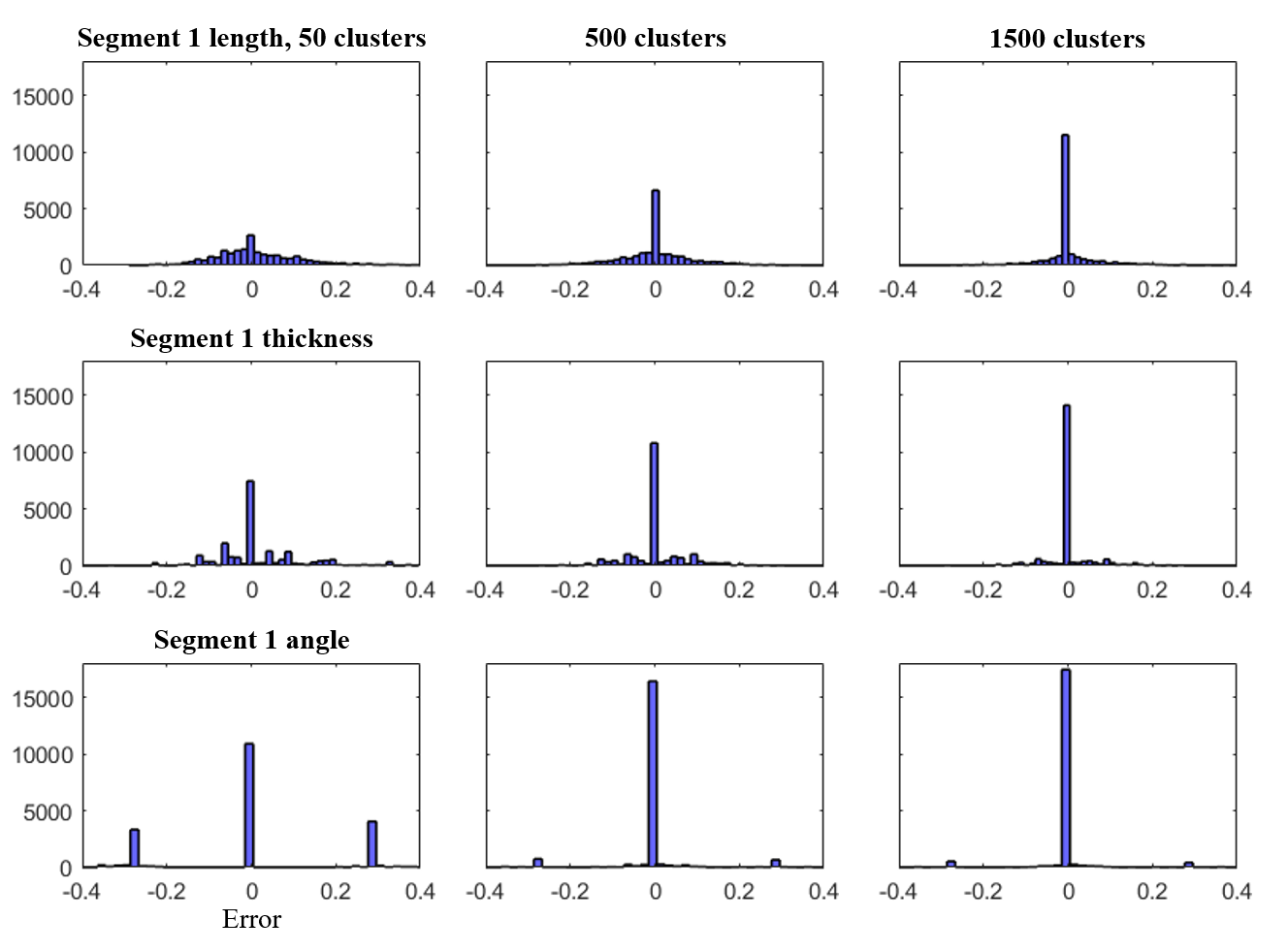}
\vspace{-.2in}
\caption{Histograms of the errors in three variables (segment 1 length along the top row, thickness in the middle row, and angle on the bottom row) for 50 (left column), 500 (middle column), and 1500 (right column) clusters.}
\label{fig:VarErrors}
\end{figure}

The $L_2$ error on its own does not inform us of whether a test set bracket is well represented by a standardized bracket, so we consider the errors on individual variables, as in Figure \ref{fig:VarErrors}. These subplots show histograms of the errors in the normalized segment 1 length, thickness, and angle at 50, 500, and 1500 clusters. The histograms are made using data from all 400 test brackets over all 50 cross validation runs, for a total of 20,000 test brackets. The three variables show different distributions: Length has an approximately normal distribution; thickness, whose unnormalized values are on a much smaller scale than length, is more flat beyond the spike at zero error; and angle has prominent spikes corresponding to $\pm90^\circ$. As expected, the distributions narrow as the number of standardized brackets increases for all three variables.

\begin{figure}
\centering
\includegraphics[width=0.9\textwidth]{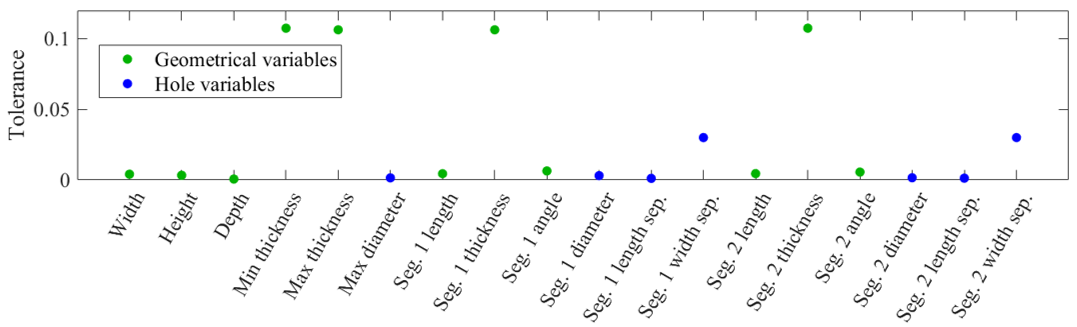}
\caption{Tolerances for the 18 non-integer variables of the data set. Geometrical variables are colored green, hole variables are blue. All tolerances are with respect to the normalized variables.}
\label{fig:Cutoffs}
\end{figure}

We now consider the number of brackets that fall below variable error thresholds for every variable, i.e. can be completely described by a standardized bracket. We assign thresholds to each numerical variable according to manufacturing tolerances: $\pm0.02$ inches for thickness, $\pm2^\circ$ for angles, $\pm0.002$ inches for hole diameter, and $\pm0.03$ inches for all other numerical variables. The normalized tolerances are shown in Fig.~\ref{fig:Cutoffs}. Thresholds of zero are applied to the one-hot encoded, integer variables. Comparing to the example variable errors shown in Fig.~\ref{fig:VarErrors}, it is apparent that these cutoffs are relatively strict.

\begin{figure}
\centering
\includegraphics[width = 0.5\textwidth]{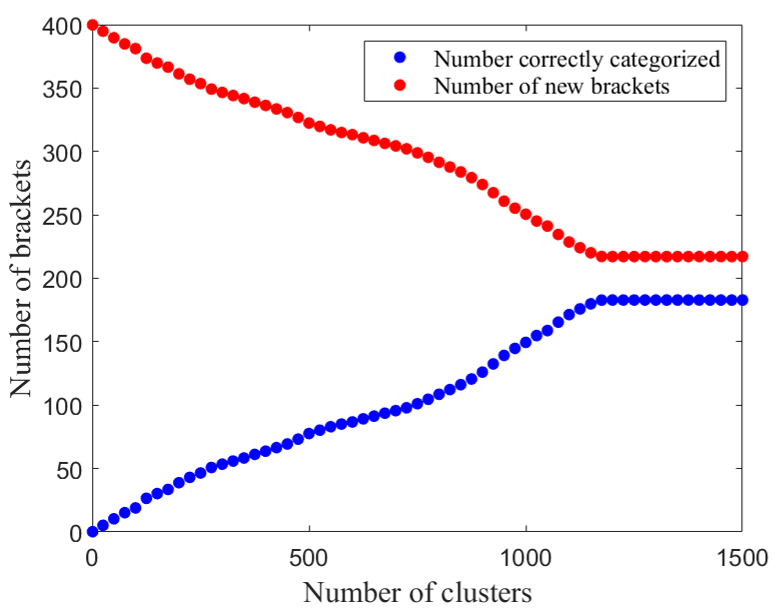}
\caption{Shown in blue are the mean number of test brackets correctly categorized. Individual thresholds are set for each variable, and a bracket is considered correctly categorized if the error in every variable is below its threshold. The red data points are the complementary set, that is, the number of brackets that cannot be described by a standardized bracket, and that therefore must be custom designed.}
\label{fig:NumCategorized}
\end{figure}

A test bracket is said to be correctly described by the nearest standardized bracket if its errors fall below the thresholds for every variable. Otherwise, the test bracket cannot be described by any standardized brackets and it must be treated as a new, unique bracket for manufacturing. Results are given in Figure~\ref{fig:NumCategorized}, where the average value of $B$, the number of correctly-described test set brackets, is plotted in blue, and the average number of new brackets needed in addition to the standardized brackets, $400 - B$, is plotted in red. Since the thresholds are fairly strict, 1500 standardized brackets correctly describes just 183 test brackets, $45\%$ of the full test set. However, the performance does not change significantly between 1200 and 1500 standardized brackets, indicating that $20\%$ of the training brackets are completely redundant.

We can improve the performance slightly by accounting for the manufacturing procedure and by improving the clustering. Clustering can be improved by incorporating fuzzy clustering techniques \citep{dunn1973fuzzy, bezdek2013pattern}, wherein brackets are allowed to belong to more than one category. We do not change the hierarchical clustering algorithm, but when considering the test set, we compare the five nearest standardized brackets, and choose one if its error falls below the thresholds in every variable, even if it is not the nearest standardized bracket in an overall $L_2$ sense.

\begin{figure}
\centering
\includegraphics[width = 0.5\textwidth]{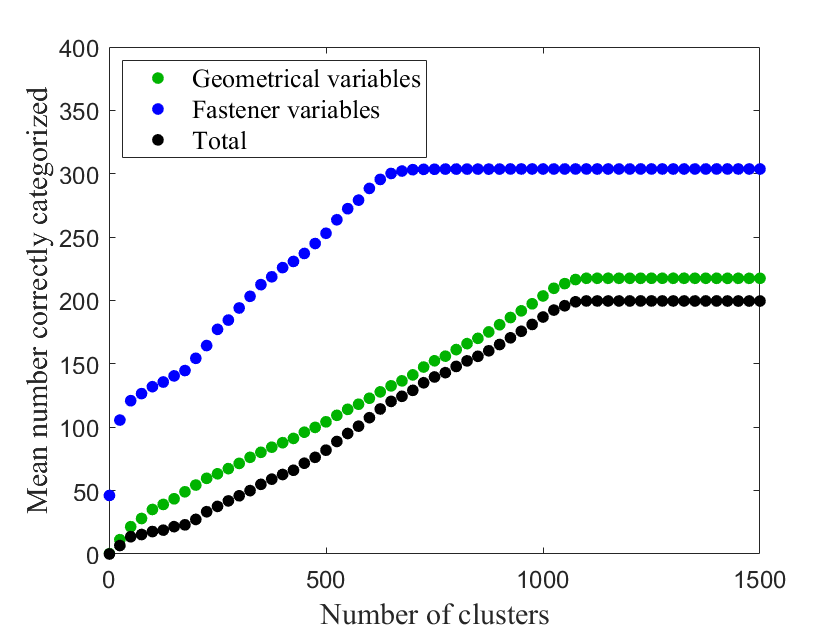}
\caption{Number correctly categorized versus number of clusters, with clustering on geometrical and fastener variables individually. The black points are the number of brackets that were correctly categorized for both groups. The same thresholds are used as in Figure \ref{fig:NumCategorized}, except that now depth is considered correct whenever the standardized bracket is too long, since the newly manufactured bracket can easily be cut down to the appropriate length. Furthermore, a fuzzy clustering formulation is used for the test brackets.}
\label{fig:Groups_NumCategorized}
\end{figure}

As for the manufacturing process, to manufacture a bracket, workers can extrude the shape, then punch in the hole pattern. Thus, there can be two independent groups of standardization, one for geometrical variables and one for fastener variables. Furthermore, it is inexpensive to extrude a bracket that is too long and cut it down to size, so the total depth can still be considered correct even if the standardized bracket is too long. Results of clustering on geometrical and hole pattern variables separately are shown in Figure \ref{fig:Groups_NumCategorized}, in green and blue, respectively. As in Figure \ref{fig:NumCategorized}, the figure shows the average number of test set brackets that are correctly categorized with thresholds on every variable, though in this case depth is considered to be correct as long as $S_{j,depth} - \sigma_{i,depth} \le \tau_{depth}$. The black data points show the brackets that were correctly categorized in both groups. With the addition of groups and fuzzy clustering, performance is improved slightly, with nearly 200 out of 400 brackets categorized correctly on average, and a plateau at about 1075 clusters. This indicates that we can reduce the number of standardized brackets by approximately 30\% while still correctly describing about 50\% of any new brackets we encounter.

\section{Conclusions and discussion}
\label{sec:conclusions}

We have considered bracket standardization in the context of aircraft manufacturing, by means of applying hierarchical clustering to a set of past brackets in order to extract a set of standardized brackets for future use. We test the performance of the standardized set by determining how many test set brackets are within some threshold of the nearest standardized bracket for every variable. We leverage the manufacturing procedure to improve performance by clustering on geometrical and hole variables independently, and find that the full data set can be reduced by about 30\% with little reduction in performance, accurately representing about 50\% of the brackets in the test set.

This was an exploration of methods, applying clustering to a large, high-dimensional data set to find a smaller set that still accurately describes it. This methodology is demonstrated on an aircraft manufacturing dataset, but the results presented here are more broadly applicable to large-scale manufacturing. We set the thresholds in order to test the performance of the reduced set of brackets, but these thresholds may be different than those encountered in a real manufacturing setting. In fact, an engineer designing a new bracket may have different tolerances depending on the situation. So the results in Section \ref{sec:results}, while suggesting that the set of past brackets may be reduced somewhat, are not exact guarantees of the performance that the standardized set will produce. Nevertheless, it is clear that bracket standardization is possible to some extent, and that it can lead to significant cost savings in aircraft manufacturing.

\subsection*{Acknowledgments} 
The authors would like to acknowledge funding support from The Boeing Company. We would also like to thanks Jim Buttrick, Tom Hogan, and Jeff Poskin for valuable discussions and support.

\bibliographystyle{elsarticle-harv}



\end{document}